# An Evolving Neuro-Fuzzy System with Online Learning/Self-learning

**Yevgeniy V. Bodyanskiy**
Kharkiv National University of Radio Electronics, Kharkiv, Ukraine,
Email: bodya@kture.kharkov.ua

**Oleksii K. Tyshchenko and Anastasiia O. Deineko**
Kharkiv National University of Radio Electronics, Kharkiv, Ukraine,
Email: {lehatish, anastasiya.deineko}@gmail.com

*Abstract*— A new neuro-fuzzy system's architecture and a learning method that adjusts its weights as well as automatically determines a number of neurons, centers' location of membership functions and the receptive field's parameters in an online mode with high processing speed is proposed in this paper. The basic idea of this approach is to tune both synaptic weights and membership functions with the help of the supervised learning and self-learning paradigms. The approach to solving the problem has to do with evolving online neuro-fuzzy systems that can process data under uncertainty conditions. The results proves the effectiveness of the developed architecture and the learning procedure.

I. INTRODUCTION

Nowadays artificial neural networks (ANNs) are widely used in Data Mining tasks, prediction tasks, identification and emulation tasks etc. under conditions of uncertainty, nonlinearity, stochasticity and chaoticity, various kinds of disturbance and noise [1-10]. They are universal approximators and are able to learn using data which characterize the object under study. If data should be processed in a sequential online mode, a convergence rate of a learning process comes to the forefront, which significantly limits the ANNs' class suitable for work under these conditions. ANNs, which use kernel activation functions (radial basis, bell-shaped, potential), are very effective from the speed optimization point of view in the learning process. Radial-basis neural networks (RBFN) are widely used, their output signal depends linearly on synaptic weights. It allows to use adaptive identification algorithms like the recurrent least-squares method, the Kaczmarz (Widrow-Hoff) algorithm etc. for their learning. However, the RBFN is exposed to the so-called "curse of dimensionality" which means that when the input space dimensionality increases, there's an exponential growth of the adjustable parameters' (weights') amount.

Neuro-fuzzy systems (NFSs) have more potential compared to neural networks [11-16], which combine learning capabilities, universal approximating properties and linguistic transparency of the results. The most popular NFSs are ANFIS and TSK-systems, whose output signal depends linearly on synaptic weights, that allows to use optimal linear identification adaptive algorithms for their learning. At the same time, to avoid gaps in the input space generated by scatter partitioning [17] which is used in ANFIS and TSK-systems, the parameters' tuning of membership functions is performed in the NFS's first hidden layer. The backpropagation algorithm is used for this purpose which is implemented with the help of multi-epochs learning [18]. Online tuning doesn't work in this case.

The idea of evolving computational systems is very popular nowadays with Data Mining scientists [19-26]. Both the system's architecture and the amount of adjustable parameters are growing rapidly while processing data. To control the RBFN activation functions' parameters (centers and matrix receptive fields) in an online mode, it was proposed in [27-29] to use the self-organizing Kohonen map [30], which provides these parameters' tuning in the self-learning process in an online mode. So the basic idea of this approach is to tune both synaptic weights and membership functions with the help of the supervised learning and self-learning paradigms.

The approach to solving the problem has to do with evolving online neuro-fuzzy systems that can process data under uncertainty conditions. It seems appropriate to extend this approach to an adaptive parameter tuning of membership functions in neuro-fuzzy systems.

The proposed tuning procedure of activation functions' parameters and their quantity was used in the proposed evolving neuro-fuzzy architecture. The tuning procedure works in an online mode. The proposed computational system was tested in forecasting tasks. The error was rather low (for a synthetic dataset: a training error was 0.02%, a test error was 1.5%; for a real-world dataset: a training error was 4.4%, a test error was 5.4%).

The remainder of this paper is organized as follows: Section 2 gives a neuro-fuzzy system's architecture and a learning procedure of output layer parameters. Section 3 describes membership functions' self-learning in the first hidden layer. Section 4 presents time-series forecasting with the help of the proposed neuro-fuzzy system. Conclusions and future work are given in the final section.

## II. A NEURO-FUZZY SYSTEM'S ARCHITECTURE AND A LEARNING PROCEDURE OF OUTPUT LAYER PARAMETERS

The proposed system's architecture (shown in fig.1) consists of five sequentially connected layers.

A $(n \times 1)$ – dimensional vector of input signals $x(k) = (x_1(k), x_2(k), ..., x_n(k))^T$ (here $k = 1, 2, ...$ is current discrete time) is fed to the input (zero) layer of the neuro-fuzzy system to be processed. The first hidden layer contains $nh$ ($h$ for each input) membership functions $\mu_{il}(x)$, $i = 1, 2, ..., n$; $l = 1, 2, ..., h$ and carries out the input space fuzzification. The second hidden layer provides the membership levels' aggregation calculated in the first layer and consists of $h$ multiplication blocks. The third hidden layer is a layer of synaptic weights to be defined during a learning process. The fourth layer is formed by two adders and calculates sums of the output signals of the second and third layers. And, finally, normalization is fulfilled in the fifth (output) layer, which results in an output signal $\hat{y}$ calculation.

Thus, if a vector signal $x(k)$ is fed to the NFS's input the first hidden layer elements calculate membership levels $0 \le \mu_{li}(x(k)) \le 1$, thus traditional Gaussian functions are commonly used as membership functions

$$\mu_{li}(x_i(k)) = \exp\left(-\frac{(x_i(k) - c_{li})^2}{2\sigma_i^2}\right) \quad (1)$$

where $c_{li}$, $\sigma_i$ are centers' parameters and width parameters correspondingly. It should be noticed that the preliminary data normalization on a certain interval, for example, $-1 \le x_i(k) \le 1$, simplifies calculations, since the width parameters $\sigma_i$ can be accepted equal for all the inputs, i.e. $\sigma_i = \sigma$.

Aggregated values $\prod_{i=1}^{n} \mu_{li}(x_i(k))$ are calculated in the second hidden layer, thus the Gaussian functions with the same width parameters $\sigma$ are

$$\prod_{i=1}^{n} \mu_{li}(x_i(k)) = \prod_{i=1}^{n} \exp\left(-\frac{(x_i(k) - c_{li})^2}{2\sigma^2}\right) = \exp\left(-\frac{\|x(k) - c_l\|^2}{2\sigma^2}\right) \quad (2)$$

(here $c_l = (c_{l1}, c_{l2}, ..., c_{ln})^T$), i.e. signals at the outputs of the multiplication blocks of the second hidden layer are similar to the signals at the neurons' outputs of the RBFN's first hidden layer.

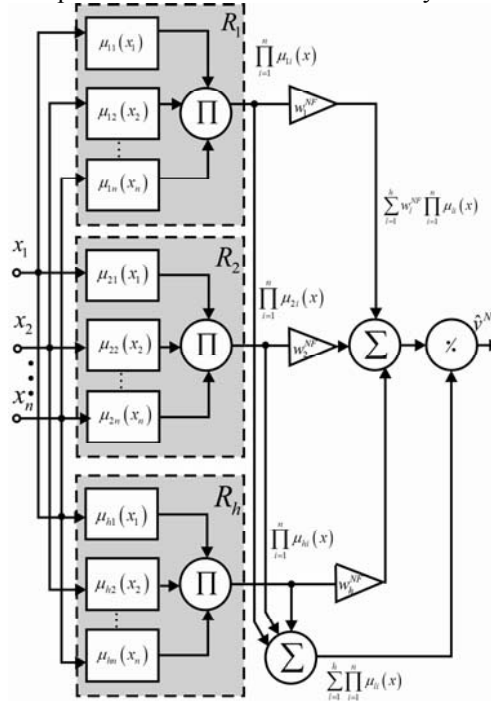

Fig.1. The neuro-fuzzy system's architecture.

The third hidden layer outputs are $w_l \prod_{i=1}^{n} \mu_{li}(x_i(k))$ (here $w_l$, $l = 1, 2, ..., h$ are synaptic weights to be defined), the fourth hidden layer outputs are $\sum_{l=1}^{h} w_l \prod_{i=1}^{n} \mu_{li}(x_i(k))$ and $\sum_{l=1}^{h} \prod_{i=1}^{n} \mu_{li}(x_i(k))$ and, finally, the system's output (the fifth layer) is

$$\hat{y}(x(k)) = \frac{\sum_{l=1}^{h} w_l \prod_{i=1}^{n} \mu_{li}(x_i(k))}{\sum_{l=1}^{h} \prod_{i=1}^{n} \mu_{li}(x_i(k))} = \sum_{l=1}^{h} w_l \frac{\prod_{i=1}^{n} \mu_{li}(x_i(k))}{\sum_{l=1}^{h} \prod_{i=1}^{n} \mu_{li}(x_i(k))} = \sum_{l=1}^{h} w_l \varphi_l(x(k)) = w^T \varphi(x(k)) \quad (3)$$

where

$$\varphi_l(x(k)) = \frac{\prod_{i=1}^{n} \mu_{li}(x_i(k))}{\sum_{l=1}^{h} \prod_{i=1}^{n} \mu_{li}(x_i(k))}, \quad w = (w_1, w_2, ..., w_h)^T$$

$\varphi(x(k)) = (\varphi_1(x(k)), \varphi_2(x(k)), ..., \varphi_h(x(k)))^T$.

It's easy to notice that the proposed system implements a nonlinear mapping of the input space into a scalar output signal like the normalized RBFN [31]. The proposed system in its architecture matches the zero-order Takagi-Sugeno-Kang system, i.e. the Wang – Mendel architecture [15].

As already mentioned, to tune the NFS's synaptic weights, one can use the well-known adaptive algorithms of identification/learning like the exponentially weighted recurrent least-squares method

$$\begin{cases} w(k) = w(k-1) + \frac{P(k-1)(y(k) - w^T(k-1)y(x(k)))}{\beta + \varphi^T(x(k))P(k-1)\varphi(x(k))} \varphi(x(k)) = w(k-1) + \frac{P(k-1)(y(k) - \hat{y}(k))}{\beta + \varphi^T(x(k))P(k-1)\varphi(x(k))} \varphi(x(k)), \\ P(k) = \frac{1}{\beta}\left(P(k-1) - \frac{P(k-1)\varphi(x(k))\varphi^T(x(k))P(k-1)}{\beta + \varphi^T(x(k))P(k-1)\varphi(x(k))}\right), \\ 0 < \beta \leq 1 \end{cases} \quad (4)$$

(here $y(k)$ is a reference learning signal, $\beta$ is a forgetting parameter of outdated information), or the one-step gradient Kaczmarz-Widrow-Hoff algorithm (it's optimal in speed):

$$w(k) = w(k-1) + \frac{y(k) - w^T(k-1)\varphi(x(k))}{\|\varphi(x(k))\|^2} \varphi(x(k)), \quad (5)$$

a learning algorithm which possesses both tracking and smoothing properties [32, 33]

$$\begin{cases} w(k) = w(k-1) + p^{-1}(k)(y(k) - w^T(k-1)\varphi(x(k)))\varphi(x(k)), \\ p(k) = \beta p(k-1) + \|\varphi(x(k))\|^2, 0 \leq \beta \leq 1 \end{cases} \quad (6)$$

and similar procedures, including the well-known linear identification procedures [31].

It's interesting to note that the procedure (6) is associated with the algorithm (4) by the ratio

$$p(k) = Tr\ P(k), \quad (7)$$

when $\beta = 0$ it gets the form of the algorithm (5).

### III. MEMBERSHIP FUNCTIONS' SELF-LEARNING IN THE FIRST HIDDEN LAYER

The membership functions' tuning process in the first hidden layer can be illustrated by a two-dimensional input vector $x(k) = (x_1(k), x_2(k))^T$ and five membership functions $\mu_{li}(x_i(k))$, $l = 1, 2, 3, 4, 5$; $i = 1, 2$ at every input. In this case, the NFS contains $nh = 10$ membership functions. Centers' initial positions $c_{li}(0)$ are evenly distributed along axes $x_1$ and $x_2$, a distance between them is defined according to this relation

$$\Delta(0) = \frac{x_{i\max} - x_{i\min}}{h-1} = \frac{2}{h-1} = 0,5 \quad (8)$$

for $-1 \leq x_i \leq 1$.

This situation is illustrated in fig.2.

In the case of the multi-dimensional input vector $x(k) \in R^n$, centers $c_{li}(0)$ are evenly distributed along the hypercube axes $[-1,1]^n$.

The first vector $x_1$ is fed to the system input (in fig.2 – $x(1) = (x_1(1), x_2(2))^T$). There are centres-"winners" $c^*_{li}(0)$ on each axis, and they are the closest ones to $x_i(1)$ in the sense of a distance

$$d_{li} = |x_i(1) - c_{li}(0)|, \qquad (9)$$

i.e.

$$c^*_{li}(0) = \arg\min\{d_{1i}, d_{2i}, ..., d_{hi}\}. \qquad (10)$$

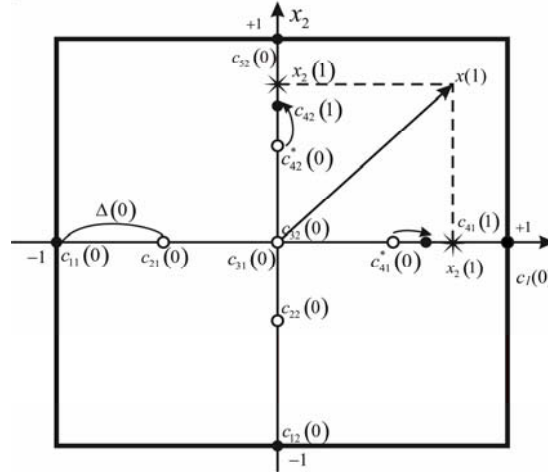

Fig. 2. Membership functions' self-learning at the first step.

Let's notice that this procedure is in fact an implementation of the competitive process according to T. Kohonen [30] but there's a slight difference that the «winner» on each axis can belong to membership functions with different indexes $l$. These «winners» in fig.2 are $c^*_{41}(0)$ and $c^*_{42}(0)$.

Then these «winners» catch up to the input signal's components $x_i(1)$ according to the Kohonen self-learning method «The winner takes all» (WTA), which can be written down for the situation in fig.2 in the form

$$c_{li}(1) = \begin{cases} c^*_{li}(0) + \eta_{li}(1)(x_i(1) - c^*_{li}(0)) \\ \qquad \text{for the winner } l = 4, \\ c_{li}(0) \text{ otherwise, } l = 1,2,3,5; \end{cases} \qquad (11)$$

and in a common case:

$$c_{li}(k) = \begin{cases} c^*_{li}(k-1) + \eta_{li}(k)(x_i(k) - c^*_{li}(k-1)) \\ \text{for the winner } l = 1,2,...,h;\ i = 1,2,...,n; \\ c_{li}(k-1) \text{ otherwise.} \end{cases} \qquad (12)$$

At the same time a value

$$\eta_{li}(k) = \frac{1}{k_{li}} \qquad (13)$$

can be accepted as a learning step parameter in the simplest case where $k_{li}$ is an amount of times when $c_{li}(k)$ was a «winner» that corresponds to the popular K-means clusterization method (stochastic approximation in the case of online processing).

In a common case, one can use an estimate which was proposed for the traditional Kohonen map [34]:

$$\begin{cases} \eta_{li}(k) = p_{li}^{-1}, \\ p_{li}(k) = \beta p_{li}(k-1) + x_i^2, \quad 0 \leq \beta \leq 1. \end{cases} \qquad (14)$$

It can be noticed that the proposed approach is a modification of the Kohonen self-learning method but the difference is a traditional self-learning procedure is implemented on the hypersphere $\|x(k)\|^2 = 1$, in our case – on the hypercube $[-1,1]^n$ (the hypersphere $\|x(k)\|^q = 1, q \to \infty$).

The combined learning/self-learning architecture of the neuro-fuzzy system is shown in fig.3.

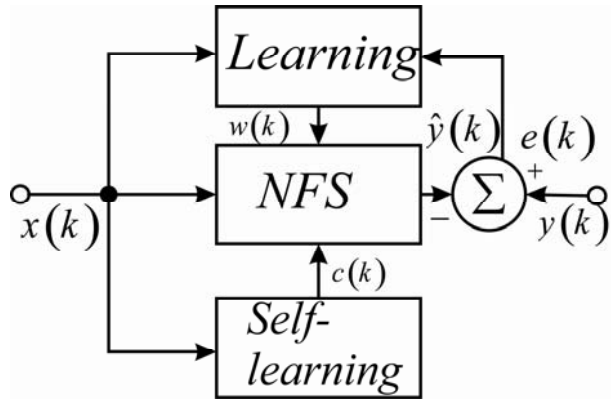

Fig.3. A combined learning/self-learning procedure.

The functioning process of the system is performed in the following way. When an input vector $x(k)$ is fed, the correction of membership functions $\mu_{li}(x_i(k))$ is carried out at first in the self-learning block, which means that centers $c_{li}(k)$ are calculated. Then the neuro-fuzzy system's output layer synaptic weights $w(k)$ are calculated on the grounds of clarified membership functions and a previously calculated synaptic weights' vector $w(k-1)$ with the help of the supervised learning algorithms ((4), (5) or (6)).

### IV. TIME-SERIES FORECASTING WITH THE HELP OF THE PROPOSED NEURO-FUZZY SYSTEM

In our experiment we used a signal generated by the Mackey-Glass equation [35] which is a non-linear differential equation

$$\frac{dx}{dt} = \beta \frac{x_\tau}{1+x_\tau^n} - \gamma x \qquad (15)$$

where $\beta, \gamma, n$ are some coefficients, $x_\tau$ is a value of a variable $x$ in the $(t-\tau)$-th time moment. The equation produces a number of periodic and chaotic values depending on parameters. In this work, these values were calculated with the help of the 4-th order Runge-Kutta method.

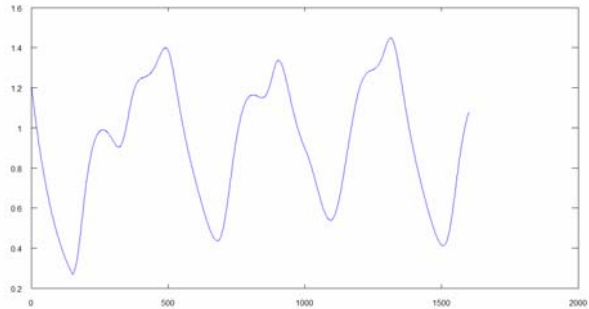

Fig.4. The Mackey-Glass time-series.

1600 values were generated with the help of the Mackey-Glass equation during a simulation procedure. These values were normalized and fed to the neuro-fuzzy system input. The sample was divided into a training set and a test set in the ratio 2:3. The learning results are shown in Fig.5:

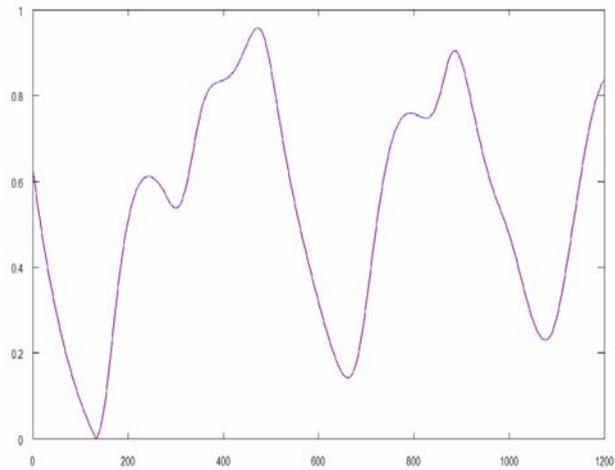
Fig. 5. The neuro-fuzzy system's learning results.

The learning error was about 0,02%. This high results' accuracy is explained by stationary properties of the Mackey-Glass time-series.

Then the system was transferred to a prediction mode. Prediction of the time-series values was fulfilled in an online mode, and elaboration of adjustable parameters in the network in a prediction mode was not carried out. The forecasting results are shown in Fig.6:

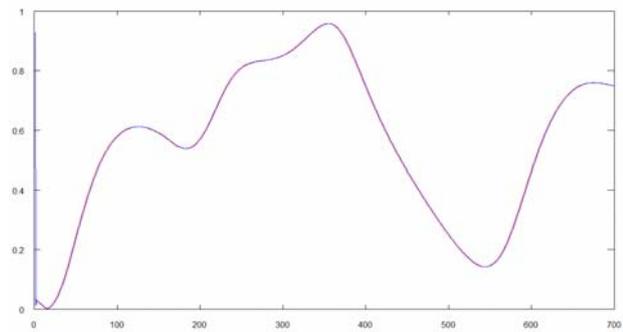
Fig.6. The prediction results of the Mackey-Glass time-series.

The error was 1.5% in a prediction mode.

To implement the experiment on real data, a sample that contains 2611 observations which characterize electricity consumption in Ukraine (December 1$^{st}$ 2008 – May 25$^{th}$ 2014, monthly) is used. To build a predictor model based on the proposed neural network's architecture data preprocessing was carried out in the form shown in Tab.1 where $x_1(k)$ is the amount of consumed electric power for the next month, $x_2(k)$ is the amount of consumed electric power in the current month, $x_3(k)$ is a seasonal component and the amount of consumed electric power 12 months ago.

Table 1. Electricity consumption in Ukraine (October 2013 – May 2014)

|  | Oct 2013 | Nov 2013 | Dec 2013 | Jan 2014 |
|---|---|---|---|---|
| $x_1(k)$ | 558275 | 543247 | 541478 | 582639 |
| $x_2(k)$ | 565640 | 558275 | 543247 | 541478 |
| $x_3(k)$ | 550135 | 534579 | 573978 | 589912 |
|  | Feb 2014 | Mar 2014 | Apr 2014 | May 2014 |
| $x_1(k)$ | 583798 | 579065 | 578932 | 575576 |
| $x_2(k)$ | 582639 | 579065 | 583798 | 578932 |

| $x_3(k)$ | 594405 | 568605 | 573412 | 598185 |

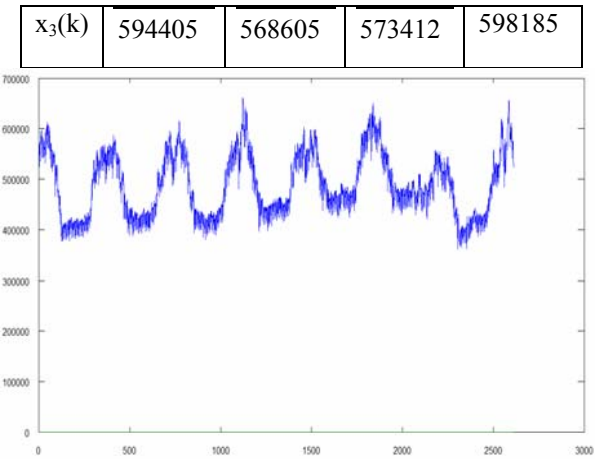
Fig.7. The graphical display of electricity consumption data.

The electricity consumption data were normalized and fed to the neuro-fuzzy network's input. The network was launched in a learning mode initially.

We obtained the results which are shown in Fig.8:

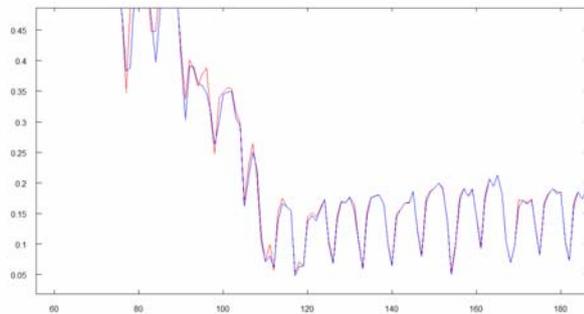
Fig.8. The neuro-fuzzy system's results in a learning mode.

The result error on a training set was 4,43%.

Then our system was launched in a prediction mode. It should be noticed that the prediction results' validation was performed in such a way: the system returned a result vector whose values were in the range [0,1] as well as the input vector's values, and then values of this output vector with the help of the quadratic error criterion were compared to actual values. Which means that the values predicted by the network itself were used as history. A number of forecasted points was limited to 14. The prediction results are shown in Fig.9:

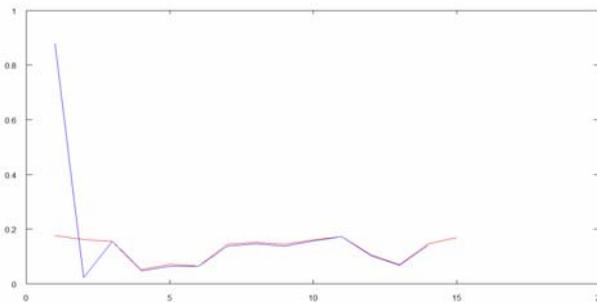
Fig.9. The prediction results of electricity consumption.

The error is 5,37%. We should notice that the total error at various time intervals was from 3 to 7%. At the beginning of the experiment centers' recalculation wasn't performed after every new value had come to the system's input, and the error was up to 20%. After the centers' values were recalculated, a function was rapidly leaving the local extremum area. This led to a sharp decrease of the result error.

## VI. CONCLUSION

The approach, which combines training of both synaptic weights and membership functions' centers and which is based on both supervised learning and self-learning, is proposed in this paper. The main advantage of the proposed approach is that it can be used in an online mode, when a training set is fed to a system's input sequentially, and its volume is not fixed beforehand. The results can be used for solving a wide class of Dynamic Data Mining problems.

ACKNOWLEDGMENT

The authors would like to thank anonymous reviewers for their careful reading of this paper and for their helpful comments.